# Comparative Analysis of Deep Learning Models for Crop Disease Detection: A Transfer Learning Approach

Saundarya Subramaniam[1], Shalini Majumdar [2], Shantanu Nadar [3], Kaustubh Kulkarni[4]

[1,2,3,4]Department of Computer Engineering, K J Somaiya School of Engineering (formerly K J Somaiya College of Engineering), Somaiya Vidyavihar University, Mumbai, India.

## Abstract

This research presents the development of an Artificial Intelligence (AI) - driven crop disease detection system designed to assist farmers in rural areas with limited resources. We aim to compare different deep learning models for a comparative analysis, focusing on their efficacy in transfer learning. By leveraging deep learning models, including EfficientNet, ResNet101, MobileNetV2, and our custom CNN, which achieved a validation accuracy of 95.76%, the system effectively classifies plant diseases. This research demonstrates the potential of transfer learning in reshaping agricultural practices, improving crop health management, and supporting sustainable farming in rural environments.

## Introduction

Artificial Intelligence is changing the way we do farming, especially when it comes to finding crop diseases early. Catching these diseases on time can stop crops from being damaged and help plants grow better. In the past, farmers had to look at plants by eye to spot problems, but this can be slow, depends on the person's judgment, and doesn't work well on big or faraway farms. Now, deep learning models—especially Convolutional Neural Networks (CNNs)—can quickly and accurately find diseases by analyzing pictures of leaves. Finding problems early is really important, because damage can happen before any signs even show up. In places like rural India, where farming is a big part of life, one bad season can cause serious money problems. This paper looks at how AI tools can help farmers by giving them fast and reliable disease predictions, even in places with fewer resources.

## Literature review

**Transfer Learning and CNN Techniques**
AI-based crop disease prediction has advanced through image classification using convolutional neural networks (CNNs). Preprocessing techniques like image resizing, normalization, and augmentation (flipping, rotation, shear, zoom) enhance model generalization and accuracy. A Tkinter-based app for real-time rice disease detection found

MobileNetV2 most accurate (95.83%), followed by DenseNet169 (91.61%) and DenseNet121 (90%) (Paneru et al., 2024). Using the Dhan-Shomadhan dataset, ResNet50 outperformed Vision Transformers (ViTs) and traditional ML models in resource-limited settings (Mehnaz & Islam, 2025). Transfer learning with pre-trained CNNs, such as MobileNet and InceptionNet, improved accuracy, with MobileNet and Random Forest achieving 92.3% (Meshram & Alvi, 2024). VGG architecture with augmentation and transfer learning reported 98.40% accuracy for grape and 95.71% for tomato leaf disease detection (Paymode & Malode, 2022). These findings underscore the effectiveness of CNNs and transfer learning in accurate, efficient disease detection.

**Datasets and Benchmarking**
High-quality datasets are crucial for improving crop disease detection models' accuracy and adaptability. The "Paddy Doctor" dataset, with 16,225 annotated images, helped a ResNet34-based model achieve a 97.50% F1-score (Petchiammal et al., 2022). The CDDM dataset, with over 137,000 images and contextual text, enables more adaptable AI models in real-world farming (Liu et al., 2025). The NUST Wheat Rust Disease dataset highlights the need for high-resolution, real-world data for accurate benchmarking (Usama et al., 2023). These challenges emphasize the need for diverse, scalable datasets to address data imbalance, environmental variability, and ensure models perform well across different crops and conditions (Khandagale et al., 2025).

**Limitations and Challenges in Crop Disease Detection**
Many agricultural datasets lack diversity, quality annotations, or class balance, which limits model performance (Guth et al., 2023). Image augmentation and transfer learning can help—accuracy improved by 25% in a ResNet-based model using just a few seed images (Xiao, 2022). Generalization is another challenge, as models trained in controlled settings often fail across different crops, regions, and conditions. Techniques like brightness and angle variation simulate real-world scenarios, improving robustness without large new datasets (Xiao, 2022).

# Methodology

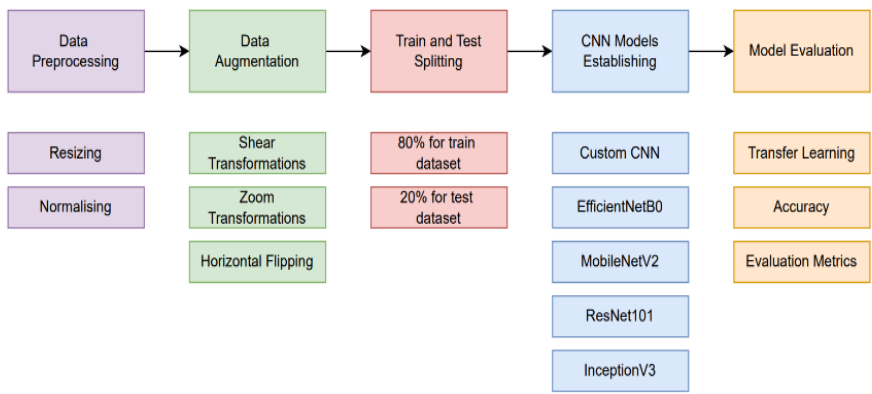

*Figure 1 Overall Workflow*

**Dataset and Image Preprocessing**

This study uses a combined dataset of over 130,000 images from the PlantVillage (54,303 images, 38 classes) and New Plant Diseases Datasets (80,000 images), enhancing species and disease diversity. Overlapping classes were unified, ensuring a variety of plant species, including apple, corn, grape, potato, and tomato, in both healthy and diseased states for effective training.

Images were resized to 150x150 pixels for the custom CNN model and 224x224 pixels for the ResNet101 model. The dataset was split into training and validation sets, with pixel values normalized. Data augmentation techniques (random shear, zoom, and flipping), were used to improve robustness and reduce overfitting.

**Custom CNN**
The custom CNN model is a sequential architecture for 150x150 RGB images. It includes four convolutional blocks, each with two Conv2D layers (3x3 kernel, ReLU, batch normalization), MaxPooling2D (2x2), and increasing dropout (0.3 to 0.5). Filter sizes double across blocks. Blocks 1-4 extract low to high-level features and deep representations. The model uses Global Average Pooling for feature aggregation and a classification head with a Dense layer (1024 units, ReLU, batch normalization, 0.5 dropout) and Softmax output for disease class prediction.

**ResNet101**
The model was trained using the Adam optimizer (learning rate $1\times10^{-3}$), categorical cross-entropy loss, and a batch size of 16 for 25 epochs, employing early stopping (patience 10 epochs). Data augmentation (random horizontal flips, shear, zoom), dropout, and batch normalization improved generalization and robustness, reducing false positives and negatives compared to transfer learning models.

ResNet101 was selected for its deep architecture and residual learning, which uses "skip connections" to solve the vanishing gradient problem, allowing effective training of deep networks. It captures both low- and high-level features in crop images, like textures and disease spots. Freezing the base model and training only the classification layers helps reduce overfitting and training time.

**InceptionV3**
InceptionV3 was chosen for its ability to capture multi-scale features using parallel convolutions (1×1, 3×3, 5×5), identifying both small details and larger patterns. Its efficient use of factorized convolutions and 1×1 bottleneck layers reduce parameters while maintaining accuracy. InceptionV3 extracts features from crop disease data by freezing early layers and fine-tuning later, making it ideal for fine-grained disease detection.

**MobileNetV2**
MobileNetV2 is a lightweight and efficient CNN developed by Google, designed for mobile and embedded vision applications. Its low memory usage and fast processing makes it perfect for resource-limited environments. By using depthwise separable convolutions and inverted residuals with linear bottlenecks, MobileNetV2 reduces parameters and computational costs while maintaining high accuracy.

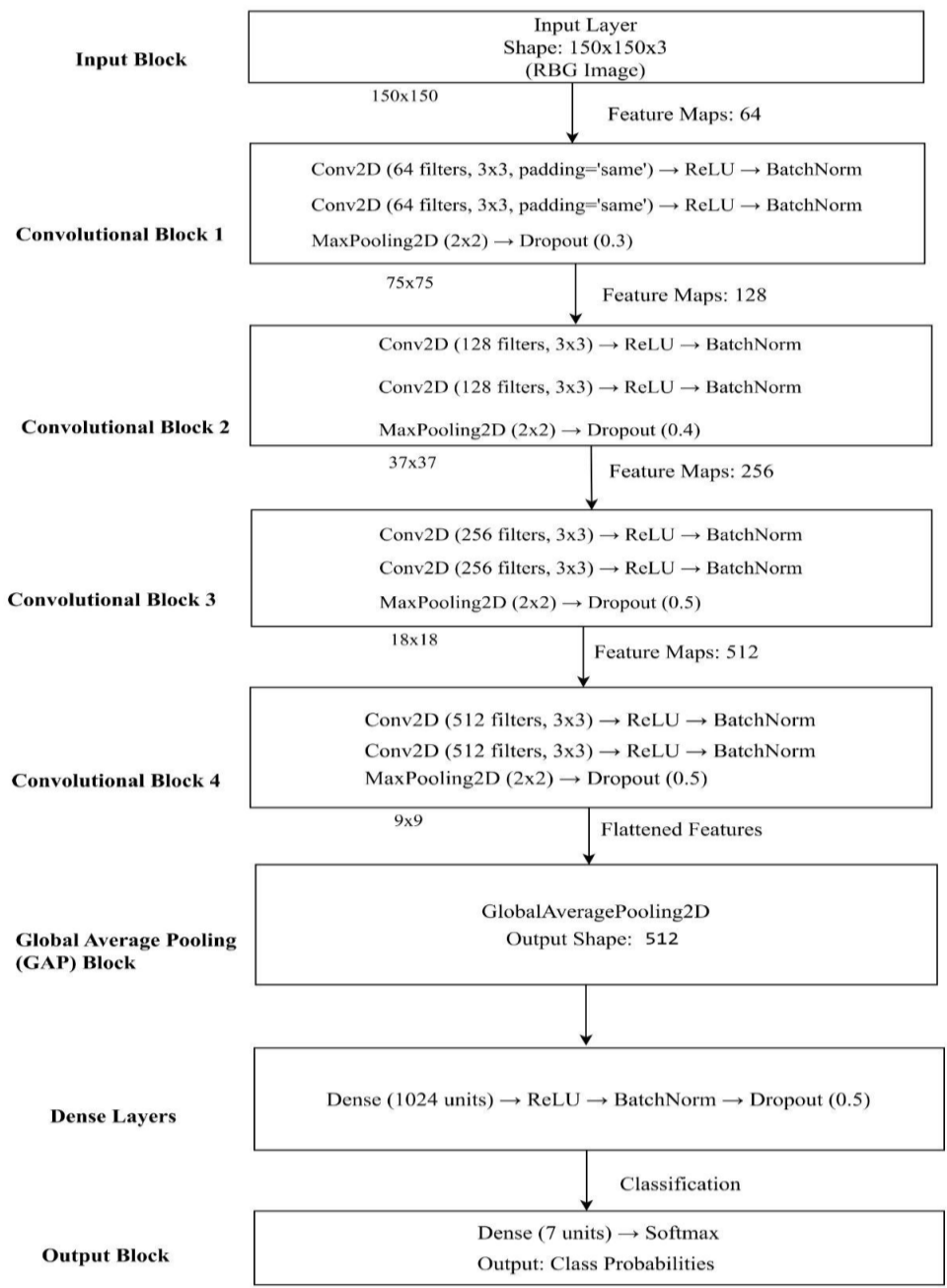

*Figure 2: Custom CNN Workflow*

**EfficientNetB0**

EfficientNetB0 was used as a lightweight, high-performing model for crop disease prediction, ideal for devices with limited resources. It uses a compound scaling method to adjust depth, width, and resolution, thus making it perfect for real-time, on-field diagnostics.

**User-Centered Deployment**
Our system aims to bridge the digital gap in Indian agriculture with a user-centered design. It includes regional language support for accessibility in different areas and an easy-to-use interface, making it simple for farmers, agronomists, and field workers with various technical skills to use. This helps them make quick, accurate decisions for managing crop health.

## Results-A comparative analysis

**Model Performance Comparison**
Table 1 summarizes the performance of four pretrained models evaluated for crop disease detection using accuracy, precision, recall, F1-score, and loss.

*Table 1: Evaluation table*

| Model | Validation Accuracy (%) | Validation Loss (%) | Precision (Macro Avg) | Recall (Macro Avg) | F1-score (Macro Avg) |
|---|---|---|---|---|---|
| Custom CNN | 95.76 | 0.96 | 0.96 | 0.96 | 0.1704 |
| ResNet101 | 96.40 | 0.95 | 0.95 | 0.95 | 0.1420 |
| EfficientNet | 97.10 | 0.97 | 0.97 | 0.97 | 0.1200 |
| MobileNet | 94.20 | 0.94 | 0.94 | 0.94 | 0.1950 |
| InceptionV3 | 96.80 | 0.96 | 0.96 | 0.96 | 0.1355 |

EfficientNet achieved the highest validation accuracy (97.10%) and top precision, recall, and F1-score (0.97). InceptionV3 followed with 96.80%, while ResNet101 (96.40%) and Custom CNN (95.76%) performed slightly lower. MobileNet had the lowest accuracy (94.20%) and highest validation loss (0.1950). Its use of depthwise separable convolutions, while efficient, likely reduced its ability to capture complex disease patterns, explaining the performance trade-off.

**Training and Validation Curve**

Training accuracy rose from 39.8% to 98.27%, while validation accuracy peaked at 99.44% and stabilized at 95.76%, indicating strong generalization. Training loss dropped from 2.39% to 0.0526%, and validation loss decreased from 6.13% to 0.1704% with minor fluctuations, confirming stable performance.

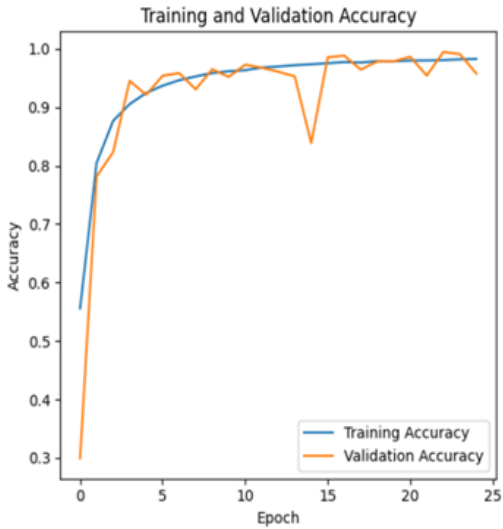

*Figure 3 Training and Validation Accuracy*

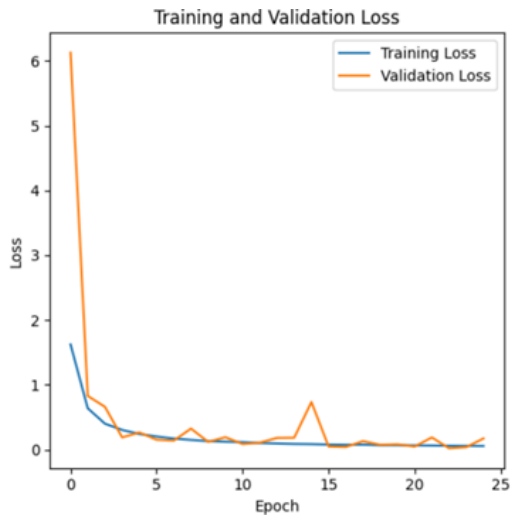

*Figure 4 Training and validation loss*

**Model Selection Justification**

EfficientNet was chosen as the final model for crop disease detection due to its top performance in accuracy, precision, recall, and F1-score. Custom CNN, with a validation accuracy of 95.76%, performed similarly to InceptionV3 and ResNet101, showing that a well-tuned custom model can also provide strong results, especially in resource-limited environments.

# Future scope

**Edge Computing and Deployment**

Optimized CNNs like MobileNetV2 and EfficientNet-lite can run on low-power devices (e.g, Raspberry Pi, NVIDIA Jetson Nano) for offline, real-time disease detection without constant internet access. Low Power Wide Area Network (LoRaWAN) enables low-bandwidth transmission of results and sensor data to servers or phones. In areas with unstable connectivity, local caching and sync mechanisms ensure reliable operation.

**Remote and Scalable Crop Monitoring**

In low-connectivity areas, farmers can transmit crop images via MMS, with compression techniques preserving key disease features. SMS fallback ensures classification results are delivered without internet access. For scalable monitoring, drones capture aerial images for real-time detection, while combining CNN outputs with satellite data enhances coverage and supports large-scale disease forecasting.

# Conclusion

In conclusion, this research demonstrates the development and deployment of an AI-powered crop disease detection system, leveraging deep learning models such as custom CNN, EfficientNet, ResNet101, and MobileNetV2. The system showed superior performance, with EfficientNet achieving the highest accuracy and efficiency. This research highlights the practical application of AI in enhancing agricultural productivity, providing timely and actionable insights for farmers in resource-limited regions.